\documentclass[sigconf]{acmart}

\usepackage{booktabs} 
\usepackage{xcolor}
\usepackage{soul}
\usepackage[utf8]{inputenc}
\usepackage{bm,color,paralist,url}
\usepackage{algorithm,algpseudocode}
\usepackage{subfig,multirow}
\usepackage{enumitem}
\usepackage{setspace}
\usepackage{hyperref}
\usepackage{amsmath}
\usepackage{amssymb}
\usepackage{graphicx}
\usepackage{balance}

\newtheorem{problem}{Problem}
\newcommand*{\Scale}[2][4]{\scalebox{#1}{$#2$}}

\copyrightyear{2018} 
\acmYear{2018} 
\setcopyright{acmcopyright}
\acmConference[MM '18]{2018 ACM Multimedia Conference}{October 22--26, 2018}{Seoul, Republic of Korea}
\acmBooktitle{2018 ACM Multimedia Conference (MM '18), October 22--26, 2018, Seoul, Republic of Korea}
\acmPrice{15.00}
\acmDOI{10.1145/3240508.3240577}
\acmISBN{978-1-4503-5665-7/18/10}

\begin{document}
\title{Causally Regularized Learning with Agnostic Data Selection Bias}

\author{Zheyan Shen}
\affiliation{%
  \institution{Tsinghua University}	
}
\email{shenzy17@mails.tsinghua.edu.cn}

\author{Peng Cui}
\affiliation{%
  \institution{Tsinghua University}	
}
\email{cuip@tsinghua.edu.cn}

\author{Kun Kuang}
\authornote{Corresponding author}
\affiliation{%
  \institution{Tsinghua University}	
}
\email{kkun2010@gmail.com}

\author{Bo Li}
\affiliation{%
  \institution{Tsinghua University}	
}
\email{libo@sem.tsinghua.edu.cn}

\author{Peixuan Chen}
\affiliation{%
  \institution{Tencent}
}
\email{noahchen@tencent.com}

\renewcommand{\shortauthors}{Z. Shen et al.}

\begin{abstract}
Most of previous machine learning algorithms are proposed based on the i.i.d. hypothesis. 
However, this ideal assumption is often violated in real applications, where selection bias may arise between training and testing process. 
Moreover, in many scenarios, the testing data is not even available during the training process, which makes the traditional methods like transfer learning infeasible due to their need on prior of test distribution. 
Therefore, how to address the agnostic selection bias for robust model learning is of paramount importance for both academic research and real applications. 
In this paper, under the assumption that causal relationships among variables are robust across domains, we incorporate causal technique into predictive modeling and propose a novel Causally Regularized Logistic Regression (CRLR) algorithm by jointly optimize global confounder balancing and weighted logistic regression.
Global confounder balancing helps to identify causal features, whose causal effect on outcome are stable across domains, then performing logistic regression on those causal features constructs a robust predictive model against the agnostic bias.
To validate the effectiveness of our CRLR algorithm, we conduct comprehensive experiments on both synthetic and real world datasets.
Experimental results clearly demonstrate that our CRLR algorithm outperforms the state-of-the-art methods, and the interpretability of our method can be fully depicted by the feature visualization.
\end{abstract}

%
%


\maketitle

\section{Introduction}
One common hypothesis in traditional machine learning is that the testing data is drawn independently from the same distribution as the training data (i.e. i.i.d hypothesis).
Then the model learned from training data can be directly applied to make predictions with the smallest empirical error on testing data.
The danger and risk caused by the violation of i.i.d. hypothesis are often being neglected in traditional machine learning methods, although those methods have made remarkable success in many difficult tasks, such as image classification, speech recognition, object localization etc. 
However, in many real applications, we can not fully control the data gathering process, then the selection bias may cause the violation of i.i.d. hypothesis.
Furthermore, in most cases the testing data is unseen during the training process, and thus the selection bias on testing data becomes agnostic.
Therefore, without considering the agnostic data selection bias, the existing predictive models are lack of robustness on different biased data, and their prediction results could be unreliable.
\begin{figure}[t]
    \centering
    \includegraphics[width=0.5\textwidth]{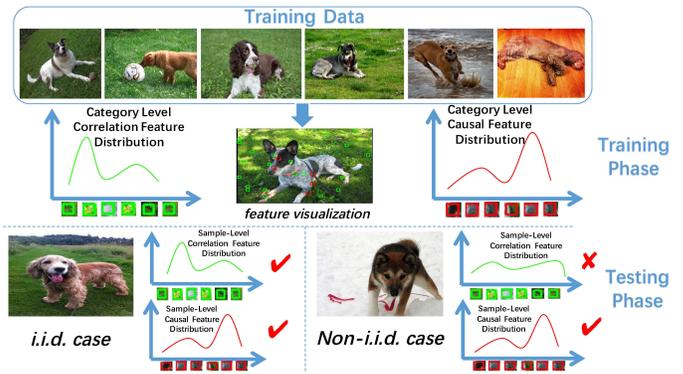}
    \caption{Illustration of the difference between correlation based and causality based methods in addressing non-i.i.d. cases.}
    \label{fig:non-i.i.d. situation}
\end{figure}
As depicted in Figure ~\ref{fig:non-i.i.d. situation}, the classifier for recognizing dogs is trained by images mostly with dogs on the grass, while tested by an image with a dog in grass context (i.e. i.i.d case) and another image with a dog in snow context (i.e. non i.i.d. case).
The correlation-based method can succeed in the i.i.d. example, but fail in the non-i.i.d. example.
The failure is mainly because the grass features are assigned with high weights in the classifier due to the fact that they are highly correlated with the label in the training set, but they do not appear in the testing image.

Recently, there are several strands of literature aiming at solving the non-i.i.d. problem induced by selection bias. 
A variety of domain adaptation methods were proposed based on feature space transformation \cite{long2015learning,fernando2013unsupervised,DBLP:conf/mm/SanginetoZRS14,DBLP:journals/pami/MaYSH14}, invariant feature learning \cite{tzeng2015simultaneous,ganin2015unsupervised} and distribution matching \cite{zadrozny2004learning,long2016deep}.
However, these methods require prior knowledge on testing data which may be unavailable in some real applications.
Domain generalization methods were proposed to overcome this dilemma, mainly based on the idea of learning a domain-agnostic model or invariant representation using training data only \cite{Li2017Deeper,Muandet2013Domain,Ghifary2015Domain}.
These methods assume the already-known selection bias(depicted by different domains) in training data and cannot generalize well to agnostic selection bias.
In this work, we investigate the learning on data with agnostic data selection bias without knowing testing data or domain information of training data. 
The targeting problem is more general than all prior work and more practical in real applications.

A reasonable way to address the agnostic selection bias is to learn a predictive model with causal variables, whose effect on outcome variable are insensitive to selection bias.
In finding these causal variables, we are much enlightened by the literature of causal inference, a powerful statistical tool for discovering causal variables and structures.
It is well recognized that causal variables are stable across different domains or data selection bias, due to the rigorous scrutiny on confounding effects \cite{rosenbaum1983central} in identifying causal variables.
The stability of causal variables is mainly reflected by the fact that the conditional distribution of outcome variable given those causal variables remains invariant across different domains.
In contrast, the correlated variables do not hold this property.
A gold standard for identifying causal effect of a variable is to conduct randomized experiments like A/B testing.
But fully randomized experiments are usually expensive and even infeasible in some scenarios.
Nevertheless, as long as the unconfoundedness assumption is satisfied \cite{rosenbaum1983central}, i.e., all confounding factors are included, and the distribution of treatment is independent of potential outcome when given the observed variables, we could precisely estimate causal effect directly from observational data.
Recently, causal inference based on observational data has become popular, and the representative methods include propensity score matching or reweighting \cite{austin2011introduction,bang2005doubly,kuang2017treatment}, markov blankets \cite{pellet2008using,koller1996toward} and confounder balancing \cite{Kuang2017Estimating,athey2016approximate,hainmueller2011entropy} etc.
However, most of these methods aim to estimate the causal effect of a variable on the output and few of them takes the advantage of causality, especially its stability across different environments in predictive modeling.

In pursuit of marrying causal analysis with non-i.i.d. learning, we still face two challenges. 
First, existing causal analysis methods are proposed in well-designed settings where typically only a small number of treatment variables are considered. 
While in high-dimensional settings of machine learning problems, we have little prior knowledge on causal relationships, and thus have to regard all variables as treatment variables. 
This makes the existing causal models infeasible due to their high computational complexity. 
Moreover, although we can first select causal variables and then learn a model based on these variables, this approach is statistically sensitive to the threshold for causal variable selection, and the step-by-step method is difficult to optimize in practice.
Hence, it is highly non-trivial to design a scalable causal learning method for prediction problems with data selection bias.

In this paper, we mainly consider the classification problem and propose a novel Causally Regularized Logistic Regression (CRLR) model for classification on data with agnostic selection bias.
The model consists of a weighted logistic loss term and a subtly designed causal regularizer. 
Specifically, the causal regularizer is designed to directly balance confounder distributions for each treatment feature through sample reweighting.
In order to reduce model complexity, we propose a global sample reweighting method which learns a common sample reweighting matrix to maximally balance confounders for all treatment features.
In this way, the weighted logistic loss and causal regularizer are jointly optimized, leading to the regression coefficients with both predictive power and causal implication. 
These merits make the resulted model be able to perform accurate and stable predictions without serious influence from agnostic selection bias.

The technical contributions of this paper are three-fold:
\begin{itemize}
  \item We investigate a new problem of learning on data with agnostic selection bias. 
  The problem setting is more general than prior work such as domain adaptation and domain generalization, and is more practical for real applications.
  \item We bring causal inference into predictive modeling and propose a novel Causally Regularized Logistic Regression model to address the above problem, where the causal regularizer and prediction loss are jointly optimized in an effective way.
  \item We conduct extensive experiments in both synthetic and real data, and the experimental results demonstrate the superiority of our method in learning on data with agnostic bias. 
  The interpretability of our method is also a notable merit.
\end{itemize}

The remaining sections are organized as follows.
Section 2 reviews the related work.
Section 3 describes the problem formulation and our CRLR algorithm.
Section 4 gives the experimental results.
Finally, Section 5 concludes the paper.

\section{Related Work}

In this section, we briefly review and discuss the previous related works which can be categorized into domain adaptation, domain generalization and causal inference.

A variety of domain adaptation methods were proposed to address the non-i.i.d. problem.
One intuition of domain adaptation is to shift the source domain distribution to align the target domain distribution and various techniques are proposed such as rejection sampling \cite{zadrozny2004learning} and bias-aware probabilistic approach \cite{liu2014robust}.
Another intuition is to learn a transformation in feature space or directly learn a domain invariant feature representation \cite{fernando2013unsupervised,long2016deep,ganin2015unsupervised,long2015learning,tzeng2015simultaneous,DBLP:conf/mm/SanginetoZRS14,DBLP:journals/pami/MaYSH14}, leveraging the powerful representation learning techniques such as deep neural networks.
A closely related task to domain adaptation is domain generalization, while testing data is unavailable during the training process.
In this setting, domain-agnostic classifiers \cite{Li2017Deeper,Ghifary2015Domain,Muandet2013Domain} are learned on multi-domain training data and applied to make prediction on unseen domains.
All the above methods require either prior knowledge on the testing data or explicit domain separation of training data, which is impractical in many real applications.
In this work, we investigate a more general and challenging problem of learning on data with agnostic selection bias, where the bias in both training and testing data is unknown.
Our targeting problem is distinct from prior works and more practical in real scenarios.

Causal inference is a powerful statistical modeling tool for explanatory analysis.
The major question in estimating causal effect is to balance the distributions of confounders across different treatment levels.
Rosenbaum and Rubin \cite{rosenbaum1983central} proposed to achieve the balance by propensity score matching or reweighting.
Methods based on propensity scores have been widely used in various fields, including economics \cite{stuart2010matching}, epidemiology \cite{funk2011doubly}, health care \cite{dos2015using}, social science \cite{lechner1999earnings} and advertising \cite{sun2015causal}.
But these methods can only handle one or a few treatment variables and cannot be directly applied in multimedia tasks in which typically a huge number of features are viewed as potential treatment variables.
There is a growing literature proposed to directly optimize sample weights to balance confounder distributions.
Hainmueller \cite{hainmueller2011entropy} introduced entropy balancing to directly adjust sample weights by the specified sample moments.
Athey et al. \cite{athey2016approximate} proposed approximate residual balancing for sample weights learning via a lasso residual regression adjustment.
Kuang et al. \cite{Kuang2017Estimating} learned a different weights of confounders and balanced confounder distributions for treatment effect estimation.
These methods provide an effective way to estimate causal effects without prior on knowledge structure, but they reweight samples targeting a single treatment variable and cannot be directly applied into predictive modeling.
We will adapt the reweighting balance technique to large-scale causal effect exploration settings we target.

\section{Causally Regularized Logistic Regression}

In this section, we provide the problem formulation, preliminaries on some key concepts of causal inference, confounder balancing, and a detailed introduction to our proposed Causally Regularized Logistic Regression (CRLR) method.

\subsection{Problem Formulation}
Here we formulate our target problem, classification on data with agnostic selection bias, as follow:

\begin{problem}[Classification on Data with Agnostic Selection Bias]
\textbf{Given} the training data $D_{train}=(\bm{X}_{train},Y_{train})$, where $\bm{X}_{train} \in \mathbb{R}^{n\times p}$ represents the features and $Y_{train} \in \mathbb{R}^{n\times 1}$ represents the label, the task is to \textbf{learn} a classifier $f_{\theta}(\cdot)$ with parameter $\theta$ to precisely predict the label of testing data $D_{test} = (\bm{X}_{test}, Y_{test})$, where $\Psi(D_{test})\neq \Psi(D_{train})$. And in the agnostic selection bias setting, we do not know how the distribution shifts from training data to unseen testing data.
\end{problem}

To solve this challenging problem, we introduce the causal inference, a powerful statistical modeling tool.
The key problem in causal inference is to estimate the causal effect of each variable on outcome or in other words, to identify the causal variable, which can be defined directly by the father nodes of outcome in Pearl's causal DAG ~\cite{Pearl2000Causality}.
When we set each variable as treatment variable to estimate its causal effect on outcome, the other variables are viewed as confounding variables.
As mentioned before, the stability of causal variables across different selection bias makes them more adequate than correlated variables in our targeting problem.
To adapt causal inference into classification problem, we regard each feature $\bm{X}_j$ as a treated variable (i.e. treatment), all the remaining features $\bm{X}_{-j} = \bm{X} \ \backslash \  \bm{X}_j$ as confounding variables (i.e. confounders), and the label $Y$ as the outcome variable. 
As we have no prior knowledge on the causal structure, it is a reasonable way to regard each variable as treatment and all the other variables as confounders \cite{hainmueller2011entropy}.
Without losing any generality, we assume all the features and labels are binary for the ease of discussion and understanding(categorical and continuous features can be converted to binary ones through binning and one-hot encoding).
Given a feature as treatment, if the feature occurs (or does not occur) in a sample, the sample becomes a treated (or control) sample.
To safely estimate the causal contribution of a given feature $\bm{X}_j$ on label $Y$, one have to remove the confounding bias induced by the different distributions of confounder $\bm{X}_{-j}$ between the treated and control groups.
After removing the confounding bias, the difference of label $Y$ between treated and control groups can be regarded as the causal contribution of feature $\bm{X}_j$ on label $Y$.

As the causal contribution $\bm{\beta} \in \mathbb{R}^{p\times 1}$ is robust and stable across different domains, we can seamlessly convert the problem of classification on data with agnostic selection bias to the following causal classification problem.

\begin{problem}[Causal Classification Problem]
\textbf{Given} the training data $D=(\bm{X},Y)$, where $\bm{X} \in \mathbb{R}^{n\times p}$ represents the features and $Y \in \mathbb{R}^{n\times 1}$ represents label, the task is to jointly \textbf{identify} the causal contribution $\bm{\beta} \in \mathbb{R}^{p\times 1}$ for all features and \textbf{learn} a classifier $f_{\beta}(\cdot)$ based on $\beta$ for classification.
\end{problem}

The key challenge in causal classification problem is how to jointly optimize the causal contribution identification and classification.
In our paper, we propose a synergistic learning algorithm composed of causal regularizer and logistic regression terms. 


\subsection{Confounder Balancing}
Here we briefly introduce some necessary background on confounder balancing which enlightens us in designing the causal regularizer.
In observational studies, confounder distributions need to be balanced to correct bias from non-random treatment assignments. 
As moments can uniquely determine a distribution, confounder balancing approaches directly balance confounder moments by adjusting weights of samples\cite{hainmueller2011entropy,athey2016approximate,Kuang2017Estimating}. 
The sample weights $W$ are learned by:
\begin{equation}
    \label{eq:ate_weight_arb1}
    {W = \arg \min_{W} \|\overline{\bm{X}}_t - \sum_{j:T_j=0}W_j\cdot \bm{X}_j\|^{2}_2.}
\end{equation}
Given a treatment feature $T$, $\overline{\bm{X}}_t$ and $\sum_{j:T_j=0}W_j\cdot \bm{X}_j$ represent the mean value of confounders on samples with and without treatment, respectively.
After confounder balancing, the correlation between a treatment variable and the output variable represents the causal effect.
Here only first-order moment (adequate for binary variable) is considered in Eq.~\ref{eq:ate_weight_arb1} and higher order moments can be easily incorporated by including more features.
Note that confounder balancing techniques are designed to estimate the causal effect of a single treatment feature. 
In our case, we need to estimate the causal effects of all features. 
This implies that we need to learn $p \times n$ sample weights, which is apparently infeasible in high-dimensional scenarios. 
Thus we propose a global balancing method as the causal regularizer in section \ref{sec:CRLR}.

\subsection{Causally Regularized Logistic Regression}
\label{sec:CRLR}
Inspired by the confounder balancing method, we propose a causal regularizer to successively set each feature as treatment variables, and find such an optimal set of sample weights that the distribution of treated and control group can be balanced for \textbf{ANY} treatment variable.
\begin{eqnarray}
    \label{causal_regularizer}
    \Scale[1.0]{\sum_{j=1}^{p} \big\|\frac{\bm{X}_{-j}^{T}\cdot(W\odot \bm{I}_j)}{W^{T}\cdot \bm{I}_j}-\frac{\bm{X}_{-j}^{T}\cdot(W\odot (1-\bm{I}_j))}{W^{T}\cdot (1-\bm{I}_j)}\big\|_2^2},
\end{eqnarray}
where $W$ is the sample weights. 
$\big\|\frac{\bm{X}_{-j}^{T}\cdot(W\odot \bm{I}_j)}{W^{T}\cdot \bm{I}_j}-\frac{\bm{X}_{-j}^{T}\cdot(W\odot (1-\bm{I}_j))}{W^{T}\cdot (1-\bm{I}_j)}\big\|_2^2$ represents the loss of confounder balancing when setting feature $j$ as treatment variable, and $\bm{X}_{-j}$ is all the remaining features (i.e. confounders), which is from $\bm{X}$ by replacing its $j^{th}$ column as $0$.
$\bm{I}_j$ means the $j^{th}$ column of $\bm{I}$, and $\bm{I}_{ij}$ refers to the treatment status of unit $i$ when setting feature $j$ as treatment variable.

Then we combine the causal regularizer and logistic regression model and propose Causally Regularized Logistical Regression (CRLR) algorithm to jointly optimize sample weights $W$ and regression coefficients $\beta$:
\begin{eqnarray}
\label{eq:causal_classification_model}
&\min& \Scale[1.0]{\sum_{i=1}^{n}W_i\cdot \log(1+\exp((1-2Y_i)\cdot (x_i\beta)))},\\
\nonumber &s.t.& \Scale[1.0]{\sum_{j=1}^{p} \big\|\frac{\bm{X}_{-j}^{T}\cdot(W\odot \bm{I}_j)}{W^{T}\cdot \bm{I}_j}-\frac{\bm{X}_{-j}^{T}\cdot(W\odot (1-\bm{I}_j))}{W^{T}\cdot (1-\bm{I}_j)}\big\|_2^2 \leq \gamma_1},\\
\nonumber   &\quad& W\succeq 0, \ \ \|W\|_2^2 \leq \gamma_2, \ \  \|\beta\|_2^2 \leq \gamma_3, \ \  \|\beta\|_1 \leq \gamma_4,\\
\nonumber &\quad& \Scale[1.0]{(\sum_{k=1}^{n}W_k-1)^{2} \leq \gamma_5},
\end{eqnarray}
where $\sum_{i=1}^{n}W_i\cdot \log(1+\exp((1-2Y_i)\cdot (x_i\beta)))$ represents the weighted logistic loss. 
Elastic net constraints $\|\beta\|_2^2\leq \gamma_3$ and $\|\beta\|_1\leq \gamma_4$ help avoid overfitting.
The term $W\succeq 0$ constrains each of sample weights to be non-negative.
With norm $\|W\|_2^2\leq \gamma_2$, we can reduce the variance of the sample weights to achieve stability.
The formula $(\sum_{k=1}^{n}W_k-1)^{2} \leq \gamma_5$ avoids all the sample weights to be $0$.

In the traditional logistic regression model, the regression coefficients capture the correlation between features and labels. 
But the highly correlated features do not imply causation due to confounding bias.  
In our model, the sample weights learned from the causal regularizer are capable of correcting the bias and globally balancing the distributions of treated and control group for any treatment features. 
Thus the estimated coefficients $\beta$ can imply causation and bear predictive power on labels simultaneously.

\subsection{Optimization}
The goal for optimizing the aforementioned model in Eq.~\ref{eq:causal_classification_model} is to minimize $\mathcal{J}(W, \beta)$ with constraints on parameters $W$ and $\beta$.
\begin{eqnarray}
\label{eq:objective_func}
\mathcal{J}(W, \beta)\!\!\!\! &=& \!\!\!\! \Scale[1.0]{\sum_{i=1}^{n}W_i\cdot \log(1+\exp((1-2Y_i)\cdot (x_i\beta)))} \\
\nonumber&& \!\!\!\!\!\!\!\! + \Scale[1.0]{\lambda_1 \sum_{j=1}^{p} \big\|\frac{\bm{X}_{-j}^{T}\cdot(W\odot \bm{I}_j)}{W^{T}\cdot \bm{I}_j}-\frac{\bm{X}_{-j}^{T}\cdot(W\odot (1-\bm{I}_j))}{W^{T}\cdot (1-\bm{I}_j)}\big\|_2^2}\\
\nonumber&& \!\!\!\!\!\!\!\! + \lambda_2 \|W\|_2^2 + \lambda_3\|\beta\|_2^2 + \lambda_4\|\beta\|_1 \\
\nonumber&& \!\!\!\!\!\!\!\! + \Scale[1.0]{\lambda_5 (\sum_{k=1}^{n}W_k-1)^{2}} \\
\nonumber &s.t.& W\succeq 0.
\end{eqnarray}

It is difficult to get an analytical solution for the final optimization problem in Eq.~\ref{eq:objective_func}.
We solve it with iterative optimization algorithm.
Firstly, we initialize sample weight $W$ and causal contribution $\beta$.
Then in each iteration, we first update $\beta$ by fixing $W$, and then update $W$ by fixing $\beta$.
These steps are described below:

\par \noindent \textbf{Update $\beta$:} When fixing $W$, the problem~(\ref{eq:objective_func}) is equivalent to optimize following objective function:
\begin{eqnarray}
\label{eq:objective_func_beta}
\mathcal{J}(\beta) &=& \Scale[1.0]{\sum_{i=1}^{n}W_i\cdot \log(1+\exp((1-2Y_i)\cdot (x_i\beta)))} \\
\nonumber&& + \lambda_3\|\beta\|_2^2 + \lambda_4\|\beta\|_1
\end{eqnarray}
which is a standard $\ell_1$ and $\ell_2$ norm regularized least squares problem and can be solved with any Elastic net solver.
Here, we use the proximal gradient algorithm \cite{parikh2014proximal} with proximal operator to optimize the objective function in (\ref{eq:objective_func_beta}).

\par \noindent \textbf{Update $W$:} By fixing $\beta$, we can obtain $W$ by optimizing~(\ref{eq:objective_func}). It is equivalent to optimize following objective function:
\begin{eqnarray}
\label{eq:objective_func_W}
\mathcal{J}(W)\!\!\!\! &=& \!\!\!\! \Scale[1.0]{\sum_{i=1}^{n}W_i\cdot \log(1+\exp((1-2Y_i)\cdot (x_i\beta)))} \\
\nonumber&& \!\!\!\!\!\!\!\! + \Scale[1.0]{\lambda_1 \sum_{j=1}^{p} \big\|\frac{\bm{X}_{-j}^{T}\cdot(W\odot \bm{I}_j)}{W^{T}\cdot \bm{I}_j}-\frac{\bm{X}_{-j}^{T}\cdot(W\odot (1-\bm{I}_j))}{W^{T}\cdot (1-\bm{I}_j)}\big\|_2^2}\\
\nonumber&& \!\!\!\!\!\!\!\! + \lambda_2 \|W\|_2^2 + \Scale[1.0]{\lambda_5 (\sum_{k=1}^{n}W_k-1)^{2}} \\
\nonumber &s.t.& W\succeq 0.
\end{eqnarray}

For ensuring non-negativity of $W$, we let $W = \omega \odot \omega$, where $\omega \in \mathbb{R}^{n\times 1}$ and $\odot$ refers to the Hadamard product.
Then the problem~(\ref{eq:objective_func_W}) can be reformulated as:
\begin{eqnarray}
\label{eq:objective_func_omega}
\mathcal{J}(\omega)\!\!\!\! &=& \!\!\!\! \Scale[1.0]{\sum_{i=1}^{n}(\omega_i\odot \omega_i)\cdot \log(1+\exp((1-2Y_i)\cdot (x_i\beta)))} \\
\nonumber&+& \!\!\!\!\!  \Scale[1.0]{\lambda_1 \sum_{j=1}^{p} \big\|\frac{\bm{X}_{-j}^{T}\cdot(\omega\odot \omega\odot \bm{I}_j)}{(\omega\odot \omega)^{T}\cdot \bm{I}_j}-\frac{\bm{X}_{-j}^{T}\cdot(\omega\odot \omega\odot (1-\bm{I}_j))}{(\omega\odot \omega)^{T}\cdot (1-\bm{I}_j)}\big\|_2^2}\\
\nonumber&+& \!\!\!\!\!  \lambda_2 \|\omega\odot \omega\|_2^2 + \Scale[1.0]{\lambda_5 (\sum_{k=1}^{n}\omega_k\odot \omega_k-1)^{2}}
\end{eqnarray}

The partial gradient of term $\mathcal{J}(\omega)$ with respect to $\omega$ is:
\begin{eqnarray}
\nonumber\frac{\partial \mathcal{J}(\omega)}{\partial \omega} &=& 2\omega\odot\log(1+\exp((1-2Y)\cdot(X\beta))) \\
\nonumber&+&\Scale[1.0]{\sum_{j=1}^{p} 4\cdot \big(\frac{\partial \mathcal{J}_{b}}{\partial \omega} \odot (\omega\cdot \mathbf{1}^{T})\big)^{T}\cdot \mathcal{J}_{b}}\\
\nonumber&+&4\omega\odot\omega\odot\omega + \Scale[1.0]{4\cdot\lambda_5 (\sum_{k=1}^{n}\omega_k\odot \omega_k-1)^{2}},
\end{eqnarray}
where $\mathbf{1}^{T} = \left( {1,1, \cdots ,1} \right)^p$, and
\begin{eqnarray}
\nonumber \mathcal{J}_{b} \!\!\!\! &=& \!\!\!\! \frac{\bm{X}_{-j}^{T}\cdot(\omega\odot \omega\odot \bm{I}_j)}{(\omega\odot \omega)^{T}\cdot \bm{I}_j}-\frac{\bm{X}_{-j}^{T}\cdot(\omega\odot \omega\odot (1-\bm{I}_j))}{(\omega\odot \omega)^{T}\cdot (1-\bm{I}_j)},\\
\nonumber \frac{\partial \mathcal{J}_b}{\partial \omega} \!\!\!\! &=& \!\!\!\! \Scale[0.95]{\frac{\bm{X}_{-j}^{T}\odot(\bm{I}_j\cdot\mathbf{1}^{T})\cdot((\omega\odot \omega)^{T}\cdot \bm{I}_j) - \bm{X}_{-j}^{T}\cdot (\omega \odot \omega \odot \bm{I}_j)\cdot \bm{I}_j^T}{\big((\omega \odot \omega)^{T}\cdot \bm{I}_j\big)^2}} \\
\nonumber \!\!\!\! &-& \!\!\!\!\!\! \Scale[0.95]{\frac{\bm{X}_{-j}^{T}\odot((1-\bm{I}_j)\cdot\mathbf{1}^{T})\cdot((\omega\odot \omega)^{T}\cdot (1-\bm{I}_j)) - \bm{X}_{-j}^{T}\cdot (\omega \odot \omega \odot (1-\bm{I}_j))\cdot (1-\bm{I}_j)^T}{\big((\omega \odot \omega)^{T}\cdot (1-\bm{I}_j)\big)^2}}
\end{eqnarray}

Then we determine the step size $a$ with line search, update $\omega$ using gradient descent, and update $W^{(t)}$ at $t^{th}$ iteration with:
\begin{eqnarray}
\nonumber W^{(t)} = \omega^{(t)} \odot \omega^{(t)}.
\end{eqnarray}

We update $\beta$ and $W$ iteratively until the objective function~(\ref{eq:objective_func}) converges. The whole algorithm is summarized in Algorithm~\ref{alg:cwb}.

\begin{algorithm}[tbp]
\caption{{Causal Regularized Logistic Regression (CRLR)}}
\label{alg:cwb}
\begin{algorithmic}[1]
\Require
Tradeoff parameters $\lambda_1>0$, $\lambda_2>0$, $\lambda_3>0$, $\lambda_4>0$, $\lambda_5>0$, Variables Matrix $\textbf{X}$ and Outcome $Y$.
\Ensure
Causal Contribution $\beta$ and Sample Weight $W$
\State Calculate Indicator Matrix $\bm{I}$ from Variables Matrix $\bm{X}$.
\State Initialize Causal Contribution $\beta^{(0)}$, Sample Weight $W^{(0)}$
\State Calculate the current value of $\mathcal{J}(W,\beta)^{(0)} = \mathcal{J}(W^{(0)},\beta^{(0)})$ with Equation~(\ref{eq:objective_func})
\State Initialize the iteration variable $t\leftarrow 0$
\Repeat
\State $t\leftarrow t+1$
\State Update $\beta^{(t)}$ by gradient descent and fix $W^{(t-1)}$
\State Update $W^{(t)}$ by gradient descent and fix $\beta^{(t-1)}$
\State Calculate $\mathcal{J}(W,\beta)^{(t)} = \mathcal{J}(W^{(t)},\beta^{(t)})$
\Until{$\mathcal{J}(W,\beta)^{(t)}$ converges or maximum iteration is reached}\\
\Return $\beta$, $W$.
\end{algorithmic}
\end{algorithm}

\subsection{Complexity Analysis}
During the procedure of optimization, the main cost is to calculate the loss $\mathcal{J}(W,\beta)$, update causal feature weights $\beta$ and sample weights $W$.
We analyze the time complexity of each of them respectively.
For the calculation of the loss, its complexity is $O(np^2)$, where $n$ is the sample size and $p$ is the dimension of variables.
For updating $\beta$, this is standard Elastic net problem and its complexity is $O(np)$.
For updating $W$, the complexity is dominated by the step of calculating the partial gradients of function $\mathcal{J}(\omega)$ with respect to variable $\omega$. The complexity of $\frac{\partial \mathcal{J}(\omega)}{\partial \omega}$ is $O(np^2)$.

In total, the complexity of each iteration in Algorithm~\ref{alg:cwb} is $O(np^2)$.

\section{Experiments}

\subsection{Dataset}
In this paper, we use both synthetic and real world datasets to validate the effectiveness of our proposed CRLR algorithm.

\textbf{Synthetic Dataset}: We generate predictive variables $\mathbf{X} = \{\mathbf{C},\mathbf{V}\} \sim N(0,1)$ with independent Gaussian distribution, where the predictive variables X are divided into two parts: causal variables C and noisy variables V. 
To make X binary, we set $X = 1$ if $X \geq 0$, otherwise $X = 0$.
Then, we generate the outcome variable $Y$ with respect to causal variables as $Y = g(C)+N(0,\epsilon)$.
To make Y binary, we set $Y = 1$ if $Y \geq 0$, otherwise $Y = 0$.

To test the effectiveness of our algorithm on data with agnostic selection bias, we generate different biased data by varying the distribution of $P(Y|V)$. 
Specifically, we vary $P(Y|V)$ via biased sample selection with a bias rate $r\in(0,1)$. 
For each sample, we select it with probability $r$ if its noisy features V equal to outcome variable Y, otherwise we select it with probability $(1-r)$. 
Note that, $r > 0.5$ corresponds to positive correlation between Y and V, $r < 0.5$ corresponds to negative correlation, and $r = 0.5$ means V and Y are independent. 
By varying the bias rate $r$, we can generate different selection bias.

\textbf{YFCC100M \cite{thomee2016yfcc100m}} is a large scale dataset which provides 100 million images and each image contains multiple tags.
In order to simulate various non-i.i.d. situations in real world, we construct a subset of original YFCC100M, which includes 10 categories, and the images in a category are divided into 5 contexts. 
For example, in the dog category, the 5 contexts are grass, beach, car, sea, and snow.
For ease of visualization and interpretation, we use SURF \cite{bay2006surf} and Bag-of-Words \cite{csurka2004visual} as features to represent images.

\textbf{WeChat Ads} is a real online advertising dataset collected from Tencecnt WeChat App\footnote{http://www.wechat.com/en/} during September 2015.
For each advertisement, there are two types of feedbacks: “Like” and “Dislike”.  
The dataset contains user feedbacks with 14,891 Likes and 93,108 Dislikes. 
For each user, we have 56 features characterizing his/her profile including (1) demographic attributes, such as age, gender, (2) number of friends, (3) device (iOS or Android), and (4) the user settings on WeChat App.
It is easy to simulate multiple subsets with different selection bias with respect to one or more profile attributes.

\textbf{Office-Caltech dataset}
The \emph{Office-Caltech} dataset is a collection of images from four distinct domains, Amazon, DSLR, Webcam and Caltech, which on average has almost a thousand labeled images.
The \emph{Office-Caltech} dataset has been commonly used in the area of domain adaptation, due to the biases created from different data collecting process \cite{torralba2011unbiased}.

\subsection{Baselines}
Due to the absence of directly related work targeting on the same problem, we implement several classic correlation-based algorithms and a two-step causal-based algorithm  to compare with CRLR.
We implement classic Logistic Regression (LR) to be the most direct baseline as our model is based on LR.
To avoid overfitting and achieve more interpretable model, we also impose $L_1$ regularizer on Logistic Regression (LR+$\bm{L_1}$) as Lasso ~\cite{tibshirani1996regression} did.
We also compare CRLR with Support Vector Machine (SVM) with linear kernel and  Multi-layer Perceptron (MLP) with 3 hidden layers.
Furthermore, we implement a straight-forward two step solution (Two-Step) which first performs causal feature selection via confounder balancing \cite{athey2016approximate} and then apply Logistic Regression.

We tuned the parameters in our algorithm and baselines via cross validation by gird searching with validation set. 
Note that in the experiments on image classification, we omit the comparison with CNN-based image classifiers, as it is infeasible to train a CNN model from scratch with only thousands of images in our dataset. 
Meanwhile, the non-i.i.d. problem setting prohibits us from using pre-trained deep models, e.g. AlexNet, because these models are trained with millions of images covering almost all possible contexts. 
In our experiments, we aim to evaluate the methods with small or moderate scale training data where non-i.i.d. problems commonly happen.

\begin{figure*}[tb]
\centering
\subfloat[\scriptsize{Trained on bias rate $r=0.65$} \label{fig:RMSE_1000_20_65}]{
  \includegraphics[width=2.2in]{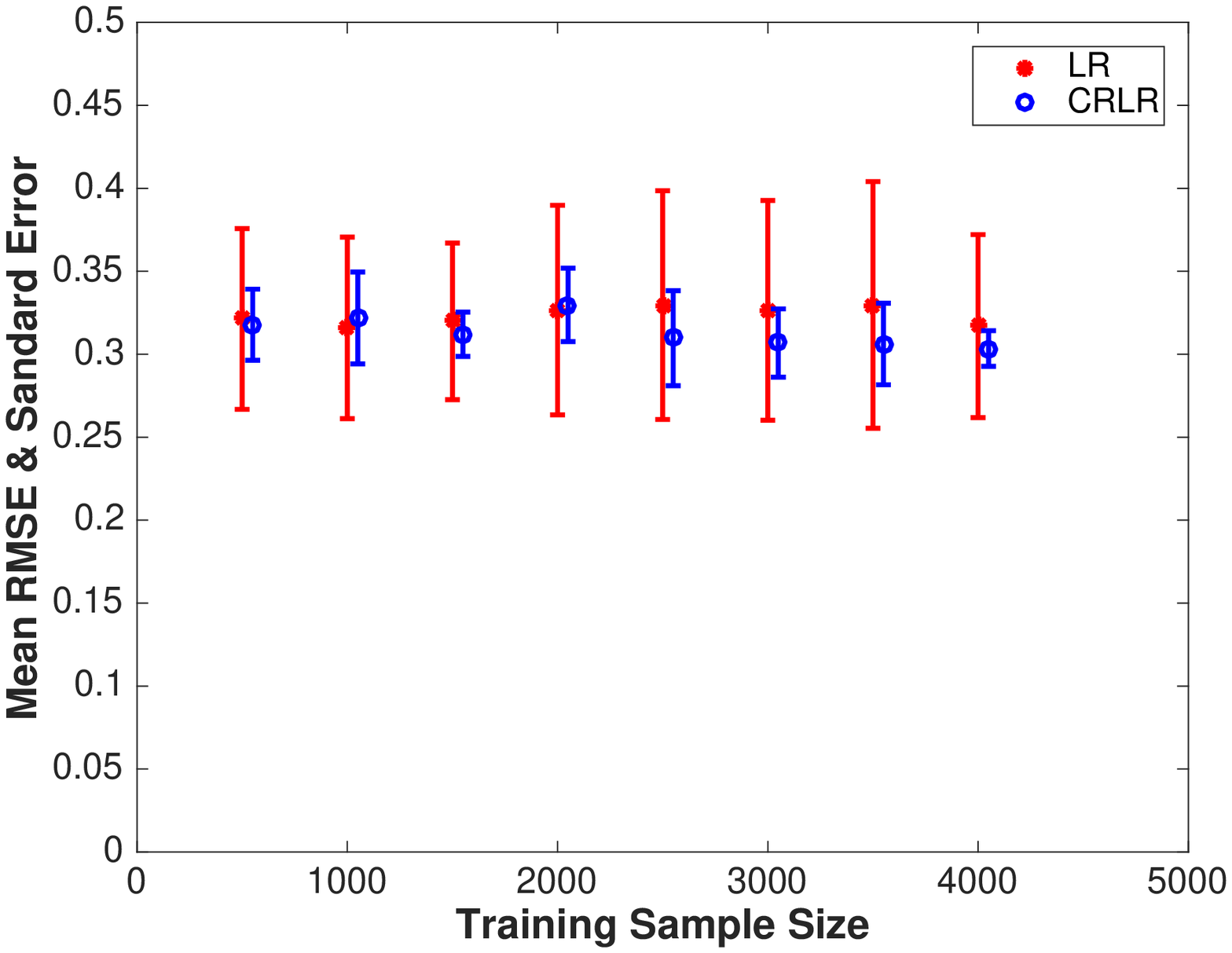}
}
\subfloat[\scriptsize{Trained on bias rate $r=0.75$}\label{fig:RMSE_2000_20_75}]{
  \includegraphics[width=2.2in]{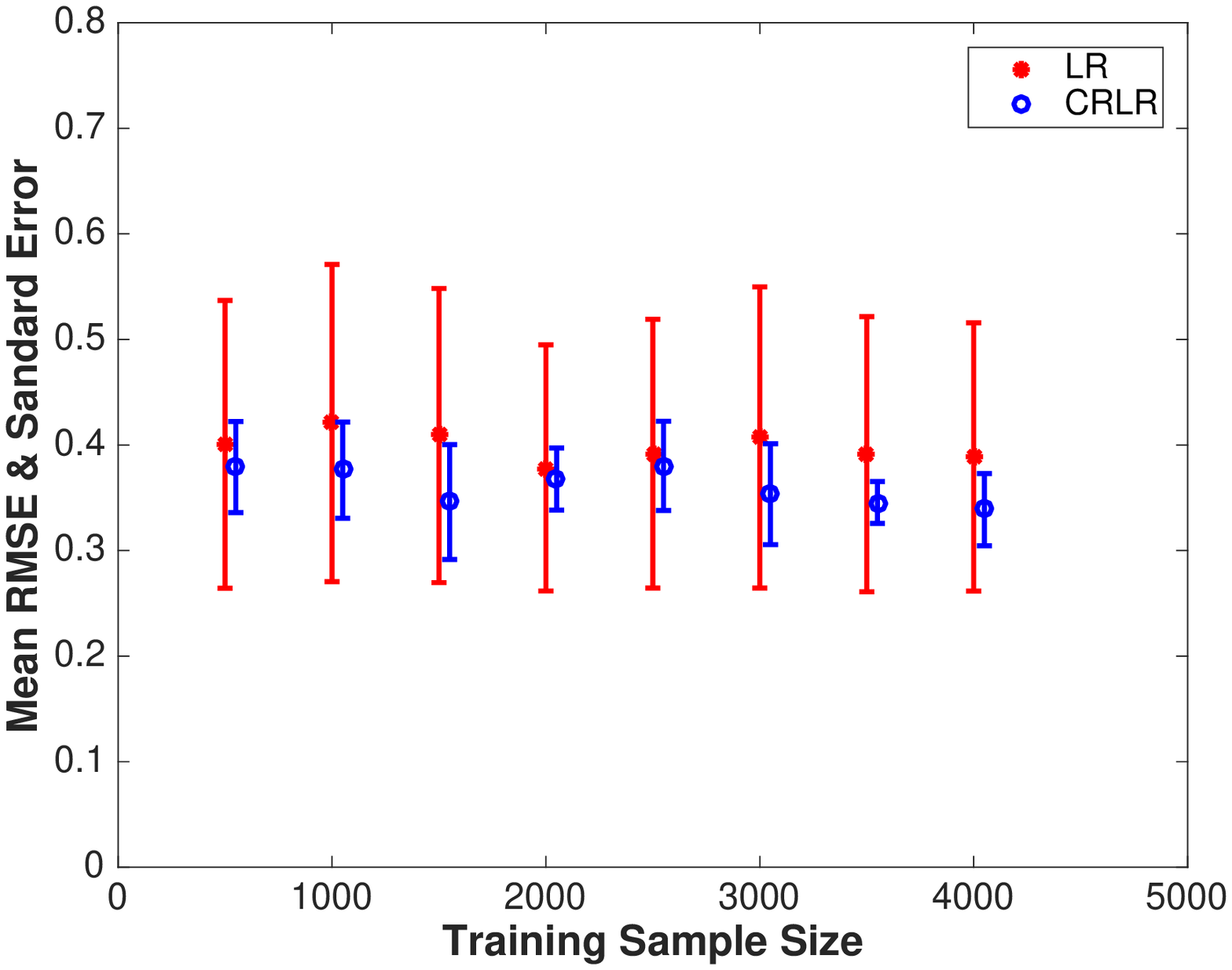}
}
\subfloat[\scriptsize{Trained on bias rate $r=0.85$}\label{fig:RMSE_4000_20_85}]{
  \includegraphics[width=2.2in]{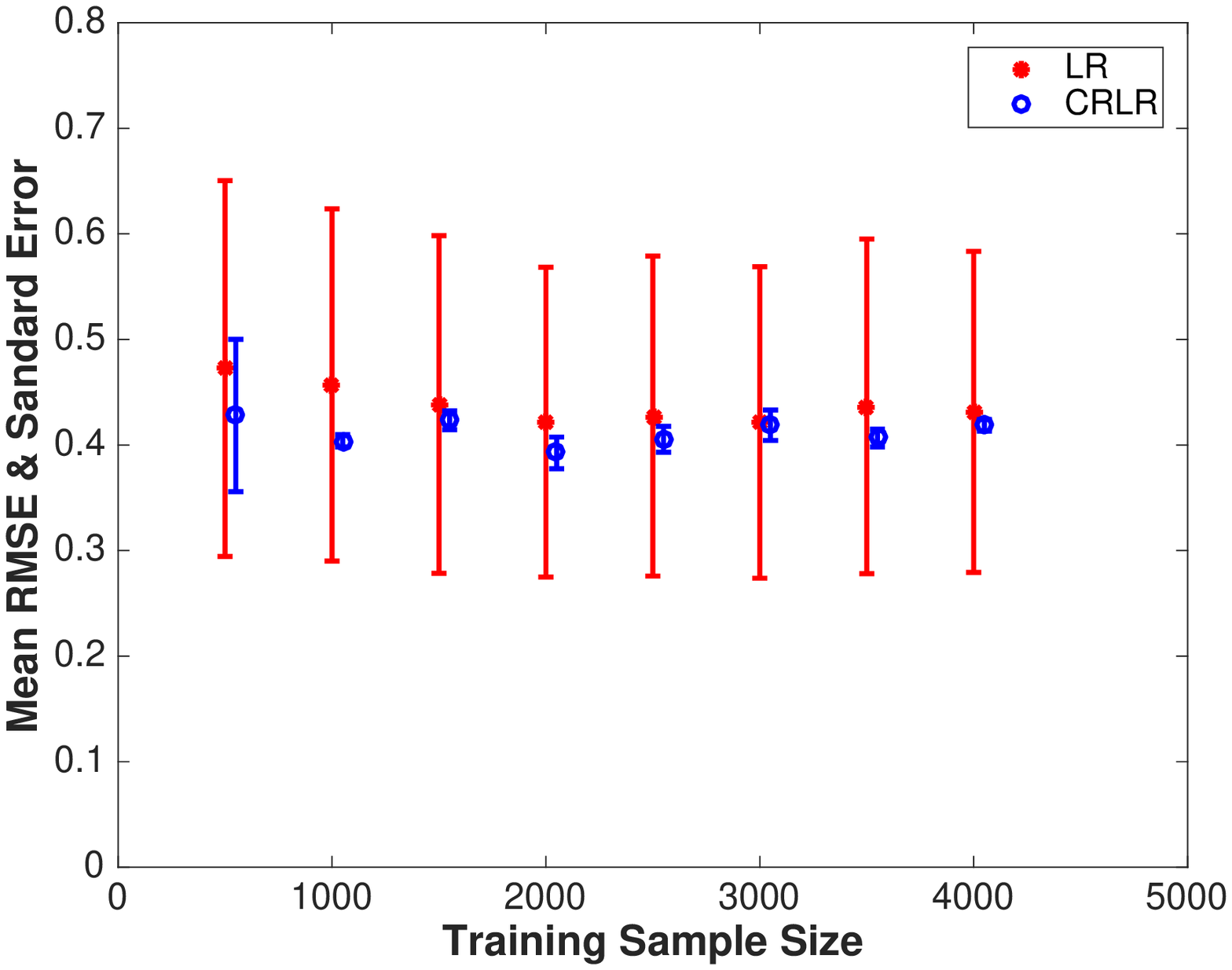}
}
\\
\caption{Average RMSE and standard Error of outcome prediction over training sample size on various training setting by varying bias rate $r$ on training set.}
\label{fig:synthetic}
\end{figure*}

\begin{table*}[htbp]
\small
\centering
\vspace{-0.1in}
{
\scalebox{0.9}{
\begin{tabular}{|c||c|c||c|c||c|c||c|c||c|c||}
\hline
&\multicolumn{2}{|c||}{bird}&\multicolumn{2}{|c||}{bridge}&\multicolumn{2}{|c||}{car}&\multicolumn{2}{|c||}{cat}&\multicolumn{2}{|c|}{church}\\
\hline
&Accuracy&F1&Accuracy&F1&Accuracy&F1&Accuracy&F1&Accuracy&F1\\
\hline
LR&0.629&0.414&\textbf{0.644}&0.450&0.709&0.588&0.617&0.456&0.760&0.637\\
\hline
LR+$L_1$&0.582&0.283&0.630&0.413&0.692&0.559&0.609&0.424&0.699&0.571\\
\hline
SVM&0.612&0.375&0.638&0.446&0.681&0.548&0.615&0.451&0.764&\textbf{0.660}\\
\hline
Two-Step&0.584&0.301&0.639&0.405&0.694&0.539&0.605&0.434&0.767&0.512\\
\hline
MLP&0.568&0.379&0.617&0.337&0.708&0.583&0.586&0.523&0.667&0.634\\
\hline
CRLR&\textbf{0.657}&\textbf{0.564}&0.617&\textbf{0.472}&\textbf{0.729}&\textbf{0.678}&\textbf{0.669}&\textbf{0.597}&\textbf{0.779}&0.633\\
\hline
&\multicolumn{2}{|c||}{dog}&\multicolumn{2}{|c||}{flower}&\multicolumn{2}{|c||}{horse}&\multicolumn{2}{|c||}{train}&\multicolumn{2}{|c|}{tree}\\
 \hline
&Accuracy&F1&Accuracy&F1&Accuracy&F1&Accuracy&F1&Accuracy&F1\\
 \hline
LR&0.565&0.370&0.734&0.635&0.580&0.362&0.592&0.398&0.732&0.618\\
\hline
LR+$L_1$&0.576&0.307&0.718&0.613&0.580&0.321&0.589&0.384&0.697&0.569\\
\hline
SVM&0.586&0.360&0.720&0.629&0.612&0.404&0.624&0.448&0.681&0.550\\
\hline
Two-Step&0.574&0.389&0.724&0.602&0.606&0.238&0.621&0.321&0.693&0.498\\
\hline
MLP&0.579&0.360&0.726&0.611&0.606&0.388&0.617&0.432&0.710&0.573\\
\hline
CRLR&\textbf{0.727}&\textbf{0.574}&\textbf{0.762}&\textbf{0.681}&\textbf{0.649}&\textbf{0.435}&\textbf{0.647}&\textbf{0.479}&\textbf{0.738}&\textbf{0.620}\\
\hline
\end{tabular}
}}
\caption{Results of classifiers under different contextual bias.}
\label{tab:results_overall_detail}
\end{table*}

\subsection{Experiments on Synthetic Data}

We test the algorithm over different bias rate (from 0.1 to 0.9) on testing set and calculate the average RMSE and standard error.
We plot the result in Figure \ref{fig:synthetic}.
From the result, we can see clearly that the performance of Logistic Regression present drastic fluctuation (larger standard error) over different bias rate on testing set across different training settings, while our proposed method achieve a relative more stable and accurate prediction result.
It is because CRLR takes advantage of stability of causal relationships and exploits causal contribution instead of correlation for prediction.

\subsection{Experiments on YFCC100M Dataset}
In this experiment, we simulate the non-i.i.d. situation by splitting different contexts into training, validation and testing set.
For each category, we use context 1,2,3 for training, context 4 for validation and context 5 for testing.
Since each category has different contexts, selection bias can vary dramatically among different categories.

Moreover, we perform a non-uniform sampling among different contexts in the training set and make the context 1/2/3 occupies 0.66/0.17/0.17 percentage respectively. 
This setting is consistent with the natural phenomena that visual concepts follow a power-law distribution \cite{clauset2009power}, indicating that only a few visual concepts are common and the rest majority are rare. 
We transfer this into visual contexts with a similar notion.

We report the performances in Accuracy and F1 in Table \ref{tab:results_overall_detail}.
From the results, we have following observations.
(1) Our CRLR model achieves the best performance in almost all categories (9/10). Since the major difference between CRLR and a standard Logistic Regression model is the causal regularizer, we can safely attribute the significant improvement to the effective confounder balancing term and its seamless joint with logistic regression model. 
(2) The performance of the two-step approach is much worse than CRLR, which clearly demonstrates the importance of jointly optimizing causal feature selection and classification. 
(3) Not surprisingly, the correlation-based classification methods do not work well in this setting, mainly because they erroneously put correlational but non-causal features in important positions, leading to their sensitivity to data selection bias.

An interesting question is to validate whether CRLR can perform much better in categories where bias is more serious. 

\begin{figure*}[tbp]
\centering
\includegraphics[width=0.75\textwidth]{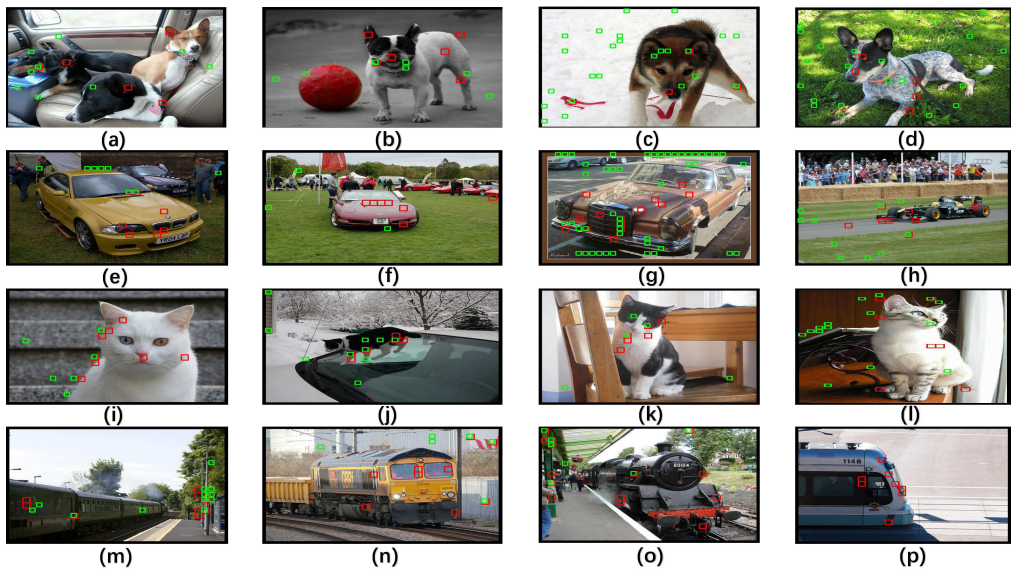}
\vspace{-0.1in}
\caption{Top 5 features selected by CRLR and Logistic Regression, the red boxes indicate the features that CRLR selects and the green boxes indicate the features that Logistic Regression selects. Note that each feature represents a visual word and may correspond to multiple bounding boxes, so the number of red and green boxes may not be equal.}
\label{fig:interpretation}
\end{figure*}

\begin{figure}[h]
\includegraphics[width=0.45\textwidth]{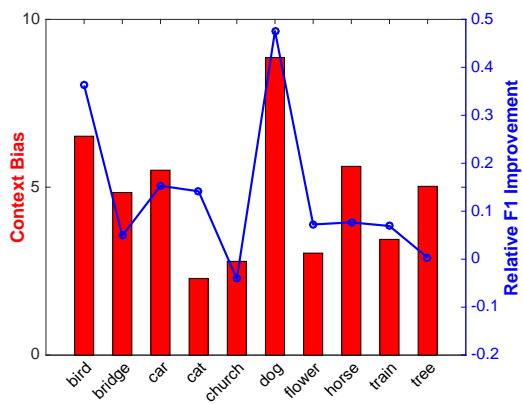}
\caption{The relationship between our CRLR algorithm performance and context bias on each category. The more context bias in data, the more relative F1 improvement of our CRLR algorithm.}
\label{fig:bias_relation}
\end{figure}
Here we quantify the bias level of a category with the EMD distance between the average feature vector of training images and the average feature vector of testing images. 
We also quantify the superiority of CRLR by its relative F1 improvement over the best baseline. 
Then we show the results in Figure \ref{fig:bias_relation}. 
We can see that relative F1 improvement and the category bias level are correlated to some degree. 
The extreme cases are more obvious. 
For example, dog category is most biased where our CRLR's relative improvement in F1 can reach about 50\%. 
In contrast, the bias in the church category is not obvious, which can account for CRLR's ordinary performance in church category in Table \ref{tab:results_overall_detail}.

A notable merit of introducing causality into predictive tasks is to make the predictive models more explainable. 
To demonstrate the interpretability of our method, we visualize the top-5 features in each category selected by CRLR and LR respectively. 
Due to space limitation, we only show some examples in 4 categories in Figure \ref{fig:interpretation}.

We can see that most of the features selected by CRLR are positioned on the major object. 
In contrast, many of the features selected by LR are context features.
From the explainable angle of view, CRLR can provide sufficient explanations on why it classifies an image into the dog category because it detects the causal features like dog nose and fur. 
We still find that our method would exploit correlation features in some cases, as depicted in Figure \ref{fig:interpretation}.(m) and \ref{fig:interpretation}.(o). 
It might because the bias level in the train category is fairly low, which weakens the effect of the causal regularizer.

\subsection{Experiments on Office-Caltech Dataset}
In this experiment, we use \emph{Office-Caltech dataset} to simulate an implicit distribution shift induced by dataset bias ~\cite{torralba2011unbiased}.
We use one domain for training and another for testing, enumerate every combination and report average accuracy.

\begin{table*}[!htbp]
    \centering
    \scalebox{0.9}{
    \begin{tabular}{|c|c|c|c|c|c|c|}
        \hline
        &LR&LR+$L_1$&SVM&Two-Step&MLP&CRLR\\
        \hline
        $a->w$&0.845&0.886&0.858&0.844&0.885&\textbf{0.897}\\
        \hline
        $a->d$&0.837&0.885&0.858&0.858&0.869&\textbf{0.887}\\
        \hline
        $w->a$&0.787&0.884&0.856&0.826&0.889&\textbf{0.900}\\
        \hline
        $w->d$&0.821&\textbf{0.901}&0.892&0.865&0.890&0.887\\
        \hline
        $d->a$&0.710&0.877&0.856&0.802&0.873&\textbf{0.900}\\
        \hline
        $d->w$&0.789&0.897&0.880&0.817&0.896&\textbf{0.898}\\
        \hline
        $a->c$&0.846&0.885&0.857&0.850&0.874&\textbf{0.895}\\
        \hline
        $w->c$&0.791&0.875&0.843&0.807&0.875&\textbf{0.896}\\
        \hline
        $d->c$&0.738&0.871&0.855&0.782&0.885&\textbf{0.897}\\
        \hline
        $c->a$&0.853&\textbf{0.904}&0.895&0.896&0.899&0.901\\        
        \hline
        $c->w$&0.859&0.896&0.887&0.886&0.888&\textbf{0.898}\\
        \hline
        $c->d$&0.841&\textbf{0.896}&0.886&0.885&0.889&0.887\\
        \hline
        mean&0.810&0.888&0.869&0.843&0.884&\textbf{0.895}\\
        \hline
    \end{tabular}  
    }
    \caption{Average accuracy on dataset bias. $a, d, w, c$ denote the four different domains Amazon, DSLR, Webcam and Caltech, respectively.}
    \label{tab:office}
\end{table*}

\textbf{Results.} From Table \ref{tab:office}, we can see that our CRLR algorithm performs the best at most of settings, showing the robustness of our algorithm even when there is no explicit distribution or domain shift.
Another interesting observation is that the advantage of our algorithm is more obvious when we have less training samples in the source domain. 
For example, the Amazon dataset is much larger than DSLR dataset, and the advantage of our algorithm over the best baseline in 'd->a' scenario is more obvious than that in 'a->d' scenario. 
This coincides with our intuition that selection bias and non-i.i.d. problems often happens when we do not have sufficient training samples and our algorithm is able to perform robust prediction in these scenarios.

\subsection{Experiments on WeChat Ads Dataset}
In this experiment, we simulate the distribution discrepancy of testing and training data by separating users into different groups according to their age.
Specifically, we split the dataset into 4 subsets by users’ age, including $Age \in [20, 30)$, $Age \in [30, 40)$, $Age \in [40, 50)$, $Age \in [50, 100)$.
And we train the baselines and CRLR on users' $Age \in [20, 30)$, and test them on all four groups.

\begin{figure}[h]
\includegraphics[width=0.45\textwidth]{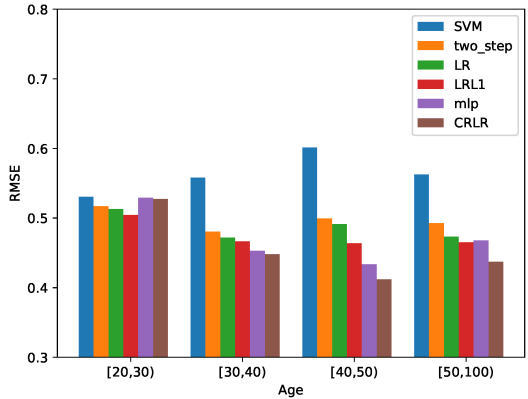}
\caption{Result of classifier under selection bias of age. All the models are trained on $Age \in [20, 30)$ and tested on all four groups.}
\label{fig:age_bias}
\end{figure}

We plot the RMSE error of each algorithm in Figure \ref{fig:age_bias}.
We can see in the group $Age \in [20,30)$ where no selection bias exists, our CRLR algorithm is comparable to baselines.
As this is a typical i.i.d. setting, the correlation between features and labels can be maximumly leveraged and most algorithms can make fairly precise predictions.
However, when tested on other three groups, where the age distributions are different from the training data, CRLR consistently outperforms the other baselines and obtains the smallest error.
It is mainly because the regression coefficients in CRLR imply causations which are more stable and insensitive to distribution shift induced by selection bias, while correlation-based methods are highly unreliable in such situations.
We also note the unsatisfactory performance of two-step method. This demonstrates the importance of jointly optimizing causal inference and predictive modeling.

\section{Conclusion and Discussion}
In this paper, we investigate a new problem of learning on data with agnostic selection bias, which is distinct from prior works and more practical in real scenarios. 
We argue that most previous methods can only preserve their predictive power when training and testing data conform to i.i.d. hypothesis or the selection bias is already known during the training process, and can not generalize well to data with agnostic selection bias.
Moreover, the results produced by those methods can hardly be interpreted and utilized for further decision making.
To address theses challenges, we introduce causality into predictive modeling and propose a novel Causally Regularized Logistic Regression (CRLR) model to jointly optimize weighted logistic loss and causal regularizer. 
We conduct comprehensive experiments on both synthetic and real world datasets and the experimental results demonstrate that our CRLR algorithm outperforms the traditional correlation-based methods in various settings. 
We also demonstrate that the top causal features selected by CRLR can provide explainable insights.

\begin{acks}
This work was supported in part by National Program on Key Basic Research Project (No. 2015CB352300), National Natural Science Foundation of China (No. 61772304, No. 61521002, No. 61531006, No. 61702296), National Natural Science Foundation of China Major Project (No.U1611461), the research fund of Tsinghua-Tencent Joint Laboratory for Internet Innovation Technology, and the Young Elite Scientist Sponsorship Program by CAST.
Bo Li's research was supported by the Tsinghua University Initiative Scientific Research Grant, No.20165080091; National Natural Science Foundation of China, No. 71490723 and No.71432004; Science Foundation of Ministry of Education of China, No. 16JJD630006.
\end{acks}

\bibliographystyle{ACM-Reference-Format}
\balance
\bibliography{sample-bibliography}

\end{document}